\documentclass[sigconf]{acmart}

\AtBeginDocument{%
  \providecommand\BibTeX{{%
    \normalfont B\kern-0.5em{\scshape i\kern-0.25em b}\kern-0.8em\TeX}}}

\setcopyright{acmcopyright}
\copyrightyear{2021}
\acmYear{2021}
\acmDOI{10.1145/1122445.1122456}

\acmConference[KDD-MLF ’21]{ML in Finance}{Aug 2021}{Virtual}

\usepackage{xcolor}
\usepackage{pgfplots}
\usepackage{tikz}
\usepackage{pgfplots}
\usepackage{pgfplotstable}
\definecolor{bblue}{HTML}{4F81BD}
\definecolor{rred}{HTML}{C0504D}
\definecolor{ggreen}{HTML}{9BBB59}
\definecolor{ppurple}{HTML}{9F4C7C}

\begin{document}

\title{Towards Theme Detection in Personal Finance Questions}


\author{John Xi Qiu}
\email{john.qiu@capitalone.com}
\affiliation{%
  \institution{Capital One}
  \city{McLean}
  \state{Virginia}
  \country{USA}
}

\author{Adam Faulkner}
\email{adam.faulkner@capitalone.com}
\affiliation{%
  \institution{Capital One}
  \city{New York}
  \state{New York}
  \country{USA}
}

\author{Aysu Ezen Can}
\email{aysu.ezencan@capitalone.com}
\affiliation{%
  \institution{Capital One}
  \city{McLean}
  \state{Virginia}
  \country{USA}
}


\begin{abstract}
Banking call centers receive millions of calls annually, with much of the information in these calls unavailable to analysts interested in tracking new and emerging call center trends.  In this study we present an approach to call center theme detection that  captures the occurrence of multiple themes in a question, using a publicly available corpus of {\it StackExchange} personal finance questions, labeled by users with topic tags, as a testbed. To capture the occurrence of multiple themes in a single question, the approach encodes and clusters at the sentence- rather than question-level. We also present a comparison of state-of-the-art sentence encoding models, including the SBERT family of sentence encoders. We frame our evaluation as a multiclass classification task and show that a simple combination of the original sentence text, Universal Sentence Encoder, and KMeans outperforms more sophisticated techniques that involve semantic parsing, SBERT-family models, and HDBSCAN.  Our highest performing  approach achieves a Micro-F1 of 0.46 for this task and we show that the resulting clusters, even when slightly noisy, contain sentences that are topically consistent with the label associated with the cluster.

\end{abstract}




\maketitle

\section{Introduction}
Call centers lie at the heart of customer-bank interactions with the typical major bank call center receiving millions of calls annually from customers. Call topics range from simple credit-card reissue requests, which trigger an associated business process, to more complex topics such as idiosyncratic hardship.  Banks typically transcribe these calls, making them available as searchable transcripts to company-internal business analysts. Alongside search, business analysts increasingly rely on the unsupervised extraction of key themes trending in call centers. Theme detection addresses a basic limitation of search, namely, its reliance on "known unknowns,"  as defined in the query. Theme detection approaches, which return sets of themes summarizing existing or emerging trends in a corpus of documents, by contrast, surface "unknown unknowns," or call center trends previously unknown to the analyst.  This information can be used to drive business decisions and agent allocation, as well as determine agent training opportunities. 

Our experimental theme detection approach employs the following pipeline approach to this task: In the first step we use a  set of simple heuristics to first identify the call sections most likely to contain a customer’s stated reason for calling.   These “Customer Problem Statements” (CPS’s)  are, in the second step, segmented into sentences or some other reduced text representation and then encoded into latent vector representations, which are then clustered into thematically distinct groupings.  Crucially, our  approach encodes and clusters sentences or reduced representations of sentences in the CPS's, rather than entire transcripts,  across an entire corpus of calls, allowing us to capture the occurrence of multiple themes in a single call. The resulting clusters are then labeled by generalizing the lexical variation found in the clustered sentences into a single, human-interpretable string.  These cluster labels are the "themes" present in a given collection of call transcripts. For example, given a cluster containing sentences in which customers, during the initial stages of the COVID crisis,  experienced flight cancelations due to travel restrictions, a theme might be {\it request\_refund\_because\_of\_Covid\_Travel\_Restrictions}.

In this study, we present the results of experiments that compare encoding and clustering techniques for the second step in the above-described pipeline. Since our data is proprietary, all reported experiments were performed using publicly available data of the same domain and roughly the same register as the CPS's in our call center data: questions posted to the personal finance site of {\it StackExchange}\footnote{https://money.stackexchange.com/}. The results of these experiments are comparable to those performed with our internal, call center data, with a combination of Universal Sentence Encoder and KMeans out-performing all other techniques.

In section 2 we review recent work specifically targeting sentence and substring encoding and clustering. In section 3, we describe the encoding models, choice of string representation, and choice of clustering evaluation metric used in our experiments. In section 4, we present experimental results and, in section 5, we summarize our results and discuss future work.

\section{Related Work}
While there has been some work\cite{2020-mlfinance} comparing embedding techniques for finance-domain text, most related work reporting the results of experimental comparisons of encoding/clustering combinations of short text involve non-finance-domain text.  Walkowiak \& Gniewkowski \cite{walkowiak2019evaluation} evaluate several embedding- and Transformer-based models in combination with different clustering and distance metrics on Polish newswire and scientific texts.  They report that a simple skip-gram model, operating over character ngrams, along with agglomerative clustering, provides the highest-quality result, where cluster quality is measured using Adjusted Mutual Information \cite{vinh2010information}. Asgari-Chenaghlu et al. \cite{asgari2020covid} track Covid-related tweets using Universal Sentence Encoder-based representations of tweets and KMeans as their clustering algorithm. Evaluation was performed on a manually annotated set of tweets, with the described approach outperforming TF-IDF- and LDA-based baselines.  In the area of event detection, the approach to detecting weather-related events across a large number of tweets, described in Arachie et al. \cite{arachie2020unsupervised}, is closest in spirit to our approach.  As with our experiments with  SRL-based representations, described in section 5, Arachie et al. first reduce tweets to “sub-events'' (i.e, noun-verb pairs) and then evaluate several encoding and clustering strategies for these sub-events. However, unlike our unsupervised approach, Arachie et al. use an ontology as a source of distance supervision, thereby restricting their sub-events to those in the scope of the ontology, and their corpus consists of tweets dealing with emerging disasters, a very different domain and register from that of bank call centers. Finally, in Chang et al. \cite{chang2021deep}, several Transformer-based encoding models are evaluated as part of a student essay answer clustering task. They report significant differences in the performance of these encoding models relative to the subject-matter of a given essay.

\begin{table*}
  \caption{The five most-frequent tags in the publicly available StackExchange training and test data and a representative StackExchange question associated with each tag.}
  \label{tab:datastats}
\begin{tabular}{p{0.08\linewidth} | p{0.04\linewidth} | p{0.68\linewidth}       }
   \toprule
     Tag    &  Freq.    & Example StackExchange Question   \\
   \midrule
   investing            &   0.088         &  "Instead of buying actual gold, bringing it into your house, then worrying about it getting stolen.  Is it possible to buy shares in gold?"  \\
   credit-card                  &  0.046  &"Is it illegal to open up a credit card account, hold it for a year, then cancel it and then turn around again and re open it and enjoy the 'first' year for free?"                \\
   mortgage                   &  0.043 & "In the UK, could a bank ask you to repay quicker a part of a mortgage (or a loan guaranteed with a property) because your risk profiles increases due to the value of property going below the loan-to-value?"            \\
   loans                  &      0.030      &   "Planning on moving and getting our house ready to sell has been more expensive than I expected.  I'd like to get this debt off of credit cards. What's the most cost effective way to do this? "    \\
   banking                   &     0.028  &  "I am opening a bank account for my company and in the registration form they ask about the director of the company, who is me.  After setting my details, they ask if I am a nominee. What is that supposed to be?"\\
 \bottomrule
\end{tabular}
\end{table*}

\section{Methodology}

\subsection{String Representation}

Our first task was to evaluate the benefit of preprocessing the sentences(s) in the question text in such a way that potentially irrelevant information is filtered out and the core semantics of the sentence is preserved.  With this goal in mind, we experimented with pre- and postprocessed predicate-argument tuples of the sentence(s) extracted from our {\it StackExchange} questions. We obtained these using a BERT-based semantic role-labeling (SRL) model \footnote{We used AllenNLP's implementation: https://allennlp.org/} \cite{shi2019simple} and utilized the following pre- and postprocessing steps: 

\begin{itemize}
\item All questions were sentence-segmented using NLTK\cite{bird2009natural}.
\item All non-personal pronouns were replaced with their single- or multiword antecedents, i.e, given the sentences  {\it I’m calling about my card}  and {\it it’s not working}, we  rewrite the latter sentence as  {\it My card not working} by resolving the {\it my card}-{\it it} coreference pair using a neural coreference model\cite{he2017deep}.\footnote{Ibid.} This allowed us to generalize many coreferring arguments in the parses.
\item All SRL parses without nominal material in their subject or object arguments were discarded.
 \item Constructions with to-infinitives, when serving as complements of {\it need} or {\it want}, such as {\it I need to know my password}, are frequent in call centers are are parsed as follows: {\it(I, need, my password)} and {\it (I, know, my password)}. The latter parse implies that the customer knows their password, which is factually incorrect. We use heuristics to identify and discard the latter parse and keep the former.
\item All arguments with adverbial or adjectival heads, such as the ARGM-ADV  argument {\it totally}, are deleted from the parse.
\item All argument text is lemmatized.\footnote{We used Spacy's lemmatization API: https://spacy.io/usage/linguistic-features\#lemmatization}
\end{itemize}

\noindent
Thus, given the sentence {\it I live in California, while my parents' estate was in Pennsylvania.}, SRL parsing and pre-and postprocessing gives us the parses {\it (i, live, in california)} and {\it (parent estate, be, in california)}, where all SRL tag information is stripped.

\subsection{Encodings}
After segmenting our question text into sentences or SRL parses, we feed each string into a pre-trained encoder. We describe experiments involving three types of encodings, each representing a different approach to encoding short text: Universal Sentence Encoder \cite{cer2017semeval}, Sentence-BERT-family models \cite{reimers2019sentence}, and, as a baseline, TF-IDF weighted ngrams.

As our first encoding type, we use the Transformer-based encoder variant of Universal Sentence Encoder (USE), as implemented in TensorFlow,\footnote{https://tfhub.dev/google/universal-sentence-encoder/4} which creates all-purpose sentence representations in two stages. First, fixed length representations of sentences are drawn from the encoding layer of the original Transformer architecture \cite{vaswani2017attention}, which captures both word ordering and context. Second, to make these representations as general as possible, a multi-task learning architecture is used to learn embeddings for multiple sentence-level tasks, such as inference and conversational input-response prediction.

Our second encoding type, Sentence-BERT (SBERT), was introduced to address BERT’s \cite{devlin2018bert} cumbersome method of determining sentence similarity, a pairwise method with a prohibitively costly time complexity. SBERT, by contrast, pools the output of BERT’s token embeddings as part of a siamese network architecture and fine-tunes on the Stanford Natural Language Inference (SNLI) and Multi-Genre Natural Language Inference (MultiNLI) corpora. The resulting, fixed-length sentence embeddings can be compared using traditional semantic similarity metrics such as cosine similarity. As with the original BERT model, better trained and smaller SBERT models--collectively called {\it SBERT-family models} herein--have been developed. These models are trained using BERT-family models such as RoBERTa \cite{liu2019roberta}, a variant of BERT optimized with longer training and more sophisticated hyperparameter tuning, and DistilRoBERTa \cite{sanh2019distilbert}, a lighter, faster version of roBERTa, as drop-in replacements for BERT.  We use three SBERT-family implementations in the experiments reported in section 5, all made available by the {\it sentence\_transformers} library\footnote{https://github.com/UKPLab/sentence-transformers}:  STSB-bert-base and STSB-Roberta-base, which were trained on SNLI and MultiNLI and then fine-tuned on a sentence similarity task, and Paraphrase-DistillRoberta-base, which was trained on millions of paraphrase examples. Finally, as our baseline encoding, we use vectors of TF-IDF-weighted ngrams.

\subsection{Clustering}
After obtaining our sentence-or-SRL-parse-level vectors, we run a clustering algorithm to aggregate semantically similar vectors. We experimented with two well established clustering approaches. First, we used K-Means clustering \cite{likas2003global} which learns to partition the embedding vector space into {\it K}-many regions, to which each sentence/SRL-parse embedding is assigned. Second, we used Hierarchical Density-Based Spatial Clustering \cite{mcinnes2017hdbscan} which learns to identify regions in embedding space with consistent densities of embedding vectors. HDBSCAN accomplishes this by first organizing instances (in our case, embeddings) into a distance weighted graph and then finds a corresponding minimum spanning tree to obtain a hierarchy of connected components comprising a cluster.

\subsection{Evaluation}

Once the sentences in the training set have been encoded and clustered, we treat the resulting clusters as a supervised classification model and evaluate the accuracy of the model on our test set using a custom evaluation procedure. 

 Let $N_{ij}$ be the number of training sentences in cluster $i$ that belong to tag $j$ and let $N_{i} = \sum_{j=1}N_{i,j} $ be the total number of sentences in cluster $i$.  We define $p_{ij} = N_{ij}/N_{i}$ as the distribution over tags for cluster $i$. So, given a 10 sentence cluster $i$ with 4 sentences tagged  {\it mortgage} and 6 sentences tagged {\it investing} our distribution  $p_{ij}$  is 0.4 for {\it mortgage}  and 0.6 for {\it investing}.

To generate a prediction for test question post $Q_{j}$, we first apply sentence tokenization, then for each sentence $S_{j} \in Q_{j}$, we find the nearest cluster $i$ by calculating the cosine distance between the encoded sentence and the cluster centroids. For the resulting set of closest clusters $U = \{ i_1 \dotso i_{K} \} $, where $K$ is the number of sentences $S_{j}$ in $Q_{j}$, each $i$ in $U$ has an associated distribution $p_{ij}$. We use these distributions to calculate a final prediction for the question by calculating 

\begin{equation}
\arg\max_{j} \frac{1}{k} \sum_{k} p_{ij}
\end{equation}

\noindent 
Thus, if a particular question has two sentences, the first of which is associated with a training cluster with the tag distribution 0.4 ({\it mortgage}) and 0.6 ({\it investing}), and the second of which is associated with a different training cluster with the tag distribution 0.3 ({\it credit-score}) and  0.7 ({\it mortgage}), the prediction for this question would be {\it mortgage} since equation (1) gives us  $(0.4+0.7)/2$ =  0.55  for {\it mortgage}, which is higher than the similarly calculated scores for {\it investing} and {\it credit-score}. Sentences can originate from a question with more than one tag though this has no impact on our tag-specific distribution calculation.

\section{Dataset}

Our main goal in dataset selection was to replicate, as much as possible, the domain and register of the CPS's in our call center data. As far as we are aware, there exist no publicly available, labeled banking call center datasets. However, the personal finance site of {\it StackExchange} makes user-generated questions, labeled with multiple topic tags, publicly available.  These questions closely resemble our customers' CPS's both in terms of domain (investments, credit-card fraud, balance transfers, etc.) and register (informal English, with a provided context followed by a question).  We collected a corpus of 32,648 personal finance {\it StackExchange} questions (all comments were excluded), timestamped between October 6th 2009 and February 27th 2021. To replicate the topics found in our call center data, we manually identified 72 tags associated with at least 50 questions in the initial dataset. These 72 tags correspond to common call reasons in our call centers (call center call reasons are manually assigned at the call level by internal quality assurance monitors as part of an ongoing business process).  We randomly split the questions using a 4:1 train-test split and removed training instances lacking one of our identified tags. Then, to simplify our experiment setting and to enable the use of multiclass classification metrics, we kept only the test questions with a single associated tag also found in our manually identified 72 tags. Our final training and test sets contain 26,118 and 2,216 questions, respectively. We checked for sampling bias when filtering out test documents with more than one tag and found that the difference in tag frequency rankings when including vs removing the multiple tagged questions was not statistically significant.

\begin{table}
\caption{Sentence counts for all 5 datasets. The number associated with each dataset indicates the max number of sentences extracted from the {\it StackExchange} questions.}
 \label{tab:sentcounts}
\begin{tikzpicture}
  \begin{axis}[
    axis x line*=bottom,
    axis y line*=none,
    every outer y axis line/.append style={draw=none},
    every y tick/.append style={draw=none},
    ymin=0,
    ymax=65000,
    xticklabel={\pgfmathtruncatemacro\tick{\tick}\tick},   
    ymajorgrids,
    y grid style={densely dotted, line cap=round},
    ylabel={Sentence Count  $\times$ 10K},
      xlabel={Dataset},
    nodes near coords,
  ]
    \addplot[
      ybar,
      draw=none,
      fill=gray,
    ] coordinates {
     ( 1,	14085)
	(2,	27725 )
	(3,	40227)
	(4,	51177)
	(5,	60404)	
    };
  \end{axis}
\end{tikzpicture}
 \end{table}

\pgfplotsset{compat=1.3}
    \pgfplotstableread{
       
x 	y1           y2
1	0.4057  0.4224
2	0.4138 0.4513
3	0.4116 0.44
4	0.4143 0.4598
5	0.4287 0.4553
1	0.2847 0.2414
2	0.3475 0.3506
3	0.356 0.366
4	0.375 0.3736
5	0.3646 0.3822

}{\loadedtable}
\begin{table}

 \caption{Preliminary experiment results. SRL-based string representations were compared to full sentences using only USE encodings.}
  \label{tab:preresults}
\begin{tikzpicture}
\node [align=center, 
    text width=3cm, inner sep=0.25cm] at (1.7cm, 3.4cm) {\textsc{USE+KMeans}};
\node [align=center,
text width=4cm, inner sep=0.25cm] at (4.9cm, 3.4cm) {\textsc{USE+HDBSCAN}};
    \begin{axis}[
        ybar,
        bar width=4pt,
        ylabel={Micro-F1},
         xlabel={Dataset},
          width  = 0.46*\textwidth,
           height = 5.2cm,
        xtick=data,
        xticklabels from table={\loadedtable}{x},
        enlargelimits=0.15,
        legend style={
            at={(0.5,-0.30)},
            anchor=north,
            legend columns=-1,
        },
        table/x expr=\coordindex,
        error bars/y dir=plus,
        error bars/y explicit,
        error bars/error bar style={
            opacity=0,
        },
        error bars/error mark options={
            rotate=90,
            mark size=0.5*\pgfplotbarwidth,
            line width=0.4pt,   
            line cap=rect,
            opacity=1,          
        },
    ]
 \addplot table [y=y1]   {\loadedtable};
        \addplot table [y=y2]   {\loadedtable};

       \draw (axis cs:4.575,0) -- ({axis cs:4.575,0}|-{rel axis cs:0.5,1});        \legend{
          SRL Parses,
           Original Sentence }
            \end{axis}
    
\end{tikzpicture}
\end{table}

 \pgfplotsset{compat=1.3}
    \pgfplotstableread{
        x       y1           y2         y3        y4        y5    
        	1	0.1557 0.4224 0.3357 0.2072 0.2205  
	2	0.1794 0.4513 0.3443 0.214 0.2217  
	3	0.2372 0.44 0.3601 0.2068 0.2329 
	4	0.2567 0.4598 0.3535 0.2253 0.2387 
	5	0.2491 0.4553 0.3581 0.2171 0.2332 
	 1      0.0715 0.2414 0.2268 0.1321 0.1666
         2      0.1131 0.3506 0.2712 0.2712 0.1758
        3        0.14300000000000002 0.366 0.2664 0.2664 0.1974
        4         0.1388 0.3736 0.2856 0.2856 0.2151
       5         0.1578 0.3822 0.2926 0.2926 0.2093

}{\loadedtable}

\begin{table*}
 \caption{Final set of experiment results comparing USE and SBERT-family encodings, using full sentences.}
\label{tab:finalresults}

\begin{tikzpicture}
\node [align=center, 
    text width=3cm, inner sep=0.25cm] at (3.90cm, 4cm) {\textsc{KMeans}};
\node [align=center,
text width=4cm, inner sep=0.25cm] at (9.65cm, 4cm) {\textsc{HDBSCAN}};
    \begin{axis}[
        ybar,
        bar width=3pt,
        ylabel={Micro-F1},
         xlabel={Dataset},
          width  = 0.85*\textwidth,
           height = 6cm,
        xtick=data,
        xticklabels from table={\loadedtable}{x},
        enlargelimits=0.15,
        legend style={
            at={(0.5,-0.30)},
            anchor=north,
            legend columns=-1,
        },
        table/x expr=\coordindex,
        error bars/y dir=plus,
        error bars/y explicit,
        error bars/error bar style={
            opacity=0,
        },
        error bars/error mark options={
            rotate=90,
            mark size=0.5*\pgfplotbarwidth,
            line width=0.4pt,   
            line cap=rect,
            opacity=1,          
        },
    ]
 \addplot table [y=y1]   {\loadedtable};
        \addplot table [y=y2]   {\loadedtable};
        \addplot table [y=y3]   {\loadedtable};
           \addplot table [y=y4]                 {\loadedtable};
              \addplot table [y=y5]                 {\loadedtable};

       \draw (axis cs:4.575,0) -- ({axis cs:4.575,0}|-{rel axis cs:0.5,1});        \legend{
            Tf-IDF,
            USE,
            DistillRoBERTa,
            RoBERTa,
            BERT }
            \end{axis}
    
\end{tikzpicture}
\end{table*}

\section{Experiments \& Results}

The {\it StackExchange} questions in our data average 7.5 sentences each, with fully 20\% of questions containing over 10 sentences.  To ensure that the clustering task is tractable, we evaluate performance for only 1 to 5 sentences from a given {\it StackExchange} question.  This means that we experimented with five separate datasets, each corresponding to the maximum number of sentences extracted from the {\it StackExchange} questions.  In Table ~\ref{tab:sentcounts}, the total sentence count for each dataset is provided.

Our experiment setup is as follows.  For each string representation (full sentences versus SRL-parses) and for each encoding (USE, SBERT, or TF-IDF) we apply encoding and clustering to each of the five datasets.  For the clustering step, we use two basic clustering techniques with tractable time complexities, KMeans and HDBSCAN.  To determine {\it k}, we used a standard variant of the elbow method: For each {\it k} starting at 100 and at $k = 100$ increments, we computed a distortion score, which is calculated as the sum of squared distances from each point in the cluster to its centroid. This procedure was performed multiple times with the most frequently occurring inflection point occurring at $ k = 700$.  We applied a similar procedure to identify HDBSCAN's optimal minimum cluster size of 5 and minimum sampling size of 3.  We then treat the resulting clusters as a classification model and evaluate the model on our 2,216 question test set (described in section 4) using the evaluation procedure described in section 3.4.


We performed two sets of experiments, the first a preliminary set of experiments evaluating the utility of comparing full sentence and SRL-parsed sentence string representations using only USE (described in section 5.1) and a second set of experiments comparing all encoding and clustering techniques using only full sentences as our string representation (described in section 5.2).

\begin{table*}
\caption{Examples of two clusters with high proportions of {\it mortgage} tags along with example text from the publicly available StackExchange dataset. The first {\it mortgage} cluster, in the column headed {\it Mostly Exclusively Mortgage Labeled} contains sentences from training set questions nearly all of which were exclusively tagged with {\it mortgage} , while the second, noisier cluster in the column {\it Mortgage Tagged with Others}, contains a sentences with a mix of {\it mortgage} and non-{\it mortgage}-related themes. }
  \label{tab:clusterexamples}
\begin{tabular}{p{0.08\linewidth} | p{0.40\linewidth}    | p{0.40\linewidth}    }
   \toprule
   
     Centrality Rank    &  Mostly Exclusively Mortgage Labeled    & Mortgage Tagged with Others  \\
   \midrule
  1      &  "I can buy a house for \$220,000 today and spend the same amount on the mortgage."     & "Since most of the early mortgage payments on my 30-yr fixed rate loan go to interest rather than principal, how do I calculate how much I need to sell for to break even?"   \\
   2     & "I would pay off the mortgage with it and then pay the loan off in place of the mortgage. "       &    "Using a rough calculator, I'd be paying about \$200/month toward the principal of the mortgage, which is more than the interest I'm accruing with my student loans."         \\
   3     &   "So my question:  Is there a way to take multiple investment property mortgages and refinance them all into a single mortgage? "  &   "I assume the interest amount I'm currently paying for my 30-year mortgage is based on a 30-year amortization trajectory."                     \\
   \bottomrule
\end{tabular}
\end{table*}

\subsection{Preliminary Experiments}

Table ~\ref{tab:preresults}  shows the results of our first, preliminary set of experiments. Since SRL parsing and the associated pre- and postprocessing adds significant overhead to our pipeline, our goal in this first set of experiments was to determine the utility of generating SRL-based representations of our {\it StackExchange} sentences using only USE as an encoder.  As is evident from Table ~\ref{tab:preresults}, for all datasets but one, the SRL-based representations underperformed relative to full sentences for both clustering techniques.  Only in dataset 1, where just a single sentence is extracted from each question, do we see competitive performance using the SRL-based representation versus its full sentence counterpart.  We also observe, for both KMeans and HDBSCAN, that, although the SRL-based encodings mostly underperform relative to full sentences, Micro-F1 scores for these encodings do increase---albeit, non-monotonically---with each, larger dataset, suggesting that increased context helps performance, even for the SRL-based encodings. Given these results, we elected to use only full sentences in our final set of experiments. 

\subsection{Final Experiments}

The results of our second, final set of experiments are given in Table ~\ref{tab:finalresults}.  For these experiments, we experimented with all encoding types and clustering algorithms using only full-sentence string representations.  The highest-performing approach was a combination of USE-based encodings of a maximum of 4 sentences from each question, and clustered using KMeans.  This combination achieved a Micro-F1 of  0.46 and beat all TF-IDF-encoding-based baselines and SBERT-family-encoding-based combinations.  All KMeans-based combinations handily beat their HDBSCAN-based counterparts, often by large margins. While the empirically determined  value for {\it  k} used by KMeans was 700, HDBSCAN tended to choose values of {\it k} as low as 100, making it impossible for the HDBSCAN-based model to capture the wider variety of topics available to the KMeans-based model.  The most surprising result was the under-performance of the SBERT-family encodings relative to USE. It's possible that fine-tuning the SBERT-family models on our data would boost performance, though the very different corpora used to train USE versus SBERT could also be a contributor to the underperformance of these models.

We can get a better sense of the quality of the clusters returned by our strongest performing  approach by examining two clusters strongly associated with one of our higher-performing tags, {\it mortgage}.   In Table~\ref{tab:clusterexamples}, we present two {\it mortgage} clusters, the first, headed {\it Mostly Exclusively Mortgage Labeled}, contains sentences originating from questions tagged almost exclusively with {\it mortgage} and the second, headed  {\it Mortgage Tagged with Others}, contains a substantial number of sentences from questions tagged with non-{\it mortgage} tags in addition to the {\it mortgage} tag. For each cluster, 3 rank-ordered sentences are presented, where rank is determined by their cosine distance to the cluster centroid.  Generally the  {\it Mostly Exclusively Mortgage Labeled} {\it mortgage}  cluster sentences tend to be the poster's original question and focus on whether or not to get a mortgage while the noisier, {\it  Mortgage Tagged with Others} cluster tends to contain sentences dealing with the context around their question---monthly payments or duration---rather than the question itself.

\begin{figure}[h]
  \centering
  \includegraphics[width=0.70\linewidth]{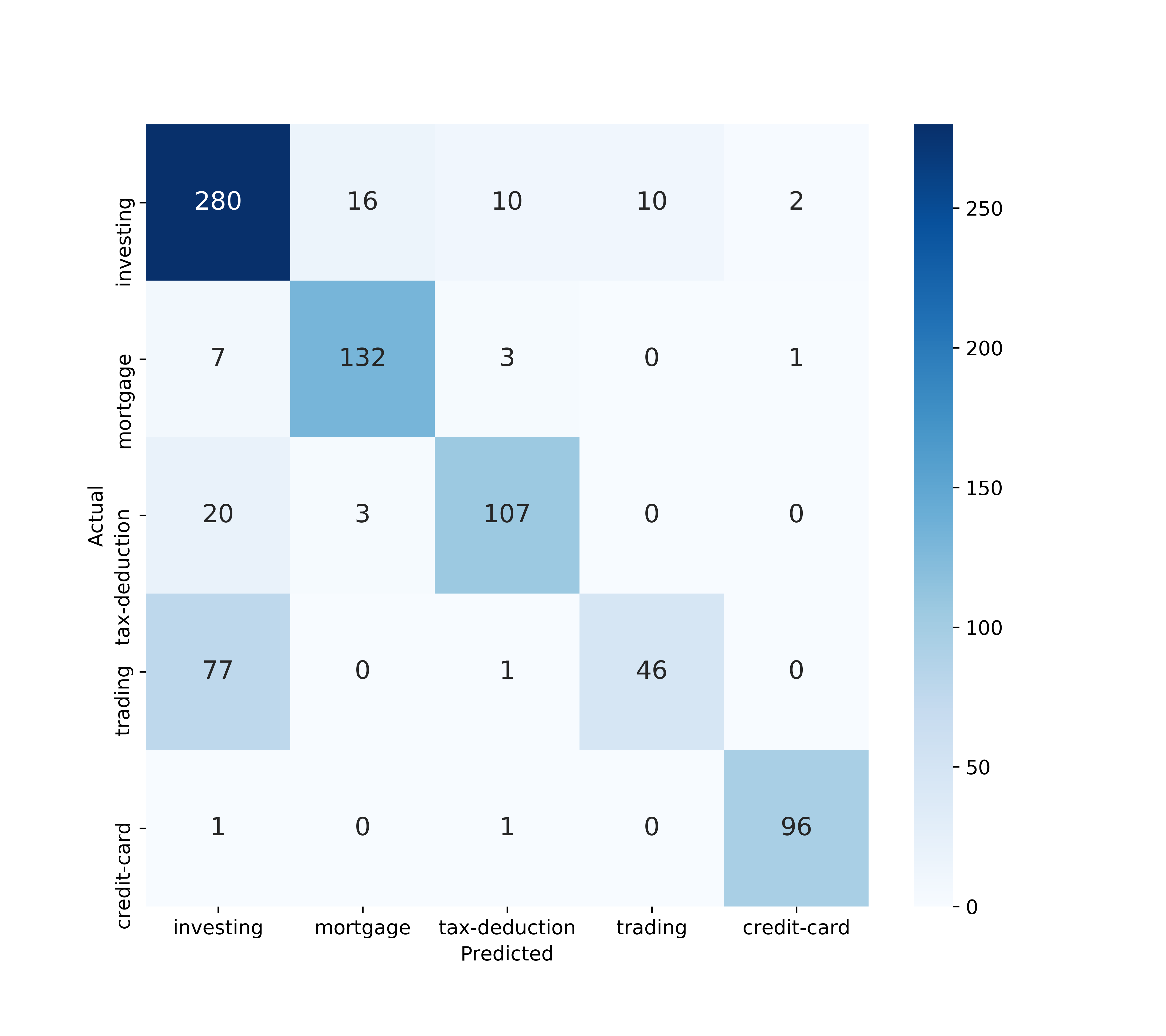}
  \caption{Confusion matrix of Top 5 most frequently labeled classes.}
\end{figure}

In our error analysis of our strongest performing model, we found that cases of misclassification were often topically related to the misclassified target tag.  Figure 1 displays a confusion matrix containing the top 5 most frequently occurring tags with predictions from our best performing experiment. For the target tag {\it trading}, we observe that a large percentage of its confused cases involved the tag {\it investing}.  Given that securities are an investment vehicle, this makes intuitive sense.  Similarly, misclassifications of {\it tax deduction} also often involve the {\it investing} tag. Again, this makes sense given the common desire to learn about tax-deductible aspects of investments such as investment interest expenses.

Another cause of the low classification performance appears to be many false positive predictions of the dominant majority class "investing." Possible solutions to this for our unsupervised approach may be to break up this ambiguous majority class into finer tags using other co-occurring tags or perhaps investigating a score normalizing scheme to mitigate the impact of the majority class.


\section{Conclusion \& Future Work}
In this study, we presented the results of a series of experiments comparing state-of-the-art sentence encoding techniques in the context of a personal finance question clustering task, using publicly available personal finance questions as a testbed. Given our interest in capturing the potential occurrence of multiple themes in a single personal finance question, our approach clusters text at the sentence- rather than question-level. Our first set of experiments compared string representations of these sentences---pre-and postprocessed SRL parses versus the original sentences---and we found that SRL-based representations underperformed relative to full sentences. A second, fuller set of experiments was presented comparing SBERT-family and USE-based encoding types. To our surprise, the state-of-the-art in sentence encoding, SBERT, underperformed relative to USE, with the strongest performing  approach consisting of a combination of USE and KMeans.

Future work will involve experimenting with the cluster labeling component. Additionally, given that our motivating use-case is to track emerging call center trends, we would like to experiment with detecting emerging trends over a given time span in our {\it StackExchange} data.


\bibliographystyle{ACM-Reference-Format}
\bibliography{kdd_mlf_21_submission}


\end{document}